%% file: main.tex
\documentclass[sigconf]{acmart}

\usepackage{multirow}

\AtBeginDocument{%
  }

\setcopyright{acmlicensed}
\copyrightyear{2018}
\acmYear{2018}
\acmDOI{XXXXXXX.XXXXXXX}

\acmConference[Conference acronym 'XX]{Make sure to enter the correct
  conference title from your rights confirmation emai}{June 03--05,
  2018}{Woodstock, NY}
\acmISBN{978-1-4503-XXXX-X/18/06}

\begin{document}

%%
%% The "title" command has an optional parameter,
%% allowing the author to define a "short title" to be used in page headers.
\title{Exploring Large Language Models for Multimodal Sentiment Analysis: Challenges, Benchmarks, and Future Directions}

%%
%% The "author" command and its associated commands are used to define
%% the authors and their affiliations.
%% Of note is the shared affiliation of the first two authors, and the
%% "authornote" and "authornotemark" commands
%% used to denote shared contribution to the research.
\author{Shezheng Song}
\email{ssz614@nudt.edu.cn}
\orcid{1234-5678-9012}
\affiliation{%
  \institution{National University of Defense Technology}
  \city{Changsha}
  \state{Hunan}
  \country{China}
}

% \author{Lars Th{\o}rv{\"a}ld}
% \affiliation{%
%   \institution{The Th{\o}rv{\"a}ld Group}
%   \city{Hekla}
%   \country{Iceland}}
% \email{larst@affiliation.org}

%%
%% By default, the full list of authors will be used in the page
%% headers. Often, this list is too long, and will overlap
%% other information printed in the page headers. This command allows
%% the author to define a more concise list
%% of authors' names for this purpose.
\renewcommand{\shortauthors}{Trovato et al.}

%%
%% The abstract is a short summary of the work to be presented in the
%% article.
\begin{abstract}
    Multimodal Aspect-Based Sentiment Analysis (MABSA) aims to extract aspect terms and their corresponding sentiment polarities from multimodal information, including text and images. While traditional supervised learning methods have shown effectiveness in this task, the adaptability of large language models (LLMs) to MABSA remains uncertain. Recent advances in LLMs, such as Llama2, LLaVA, and ChatGPT, demonstrate strong capabilities in general tasks, yet their performance in complex and fine-grained scenarios like MABSA is underexplored. In this study, we conduct a comprehensive investigation into the suitability of LLMs for MABSA. To this end, we construct a benchmark to evaluate the performance of LLMs on MABSA tasks and compare them with state-of-the-art supervised learning methods. Our experiments reveal that, while LLMs demonstrate potential in multimodal understanding, they face significant challenges in achieving satisfactory results for MABSA, particularly in terms of accuracy and inference time. Based on these findings, we discuss the limitations of current LLMs and outline directions for future research to enhance their capabilities in multimodal sentiment analysis.
\end{abstract}

%%
%% The code below is generated by the tool at http://dl.acm.org/ccs.cfm.
%% Please copy and paste the code instead of the example below.
%%


%%
%% Keywords. The author(s) should pick words that accurately describe
%% the work being presented. Separate the keywords with commas.
\keywords{Do, Not, Us, This, Code, Put, the, Correct, Terms, for,
  Your, Paper}
%% A "teaser" image appears between the author and affiliation
%% information and the body of the document, and typically spans the
%% page.

\received{20 February 2007}
\received[revised]{12 March 2009}
\received[accepted]{5 June 2009}

%%
%% This command processes the author and affiliation and title
%% information and builds the first part of the formatted document.
\maketitle

\section{Introduction}
\begin{figure}[tbp]
        \centering
        \includegraphics[width=.9\linewidth]{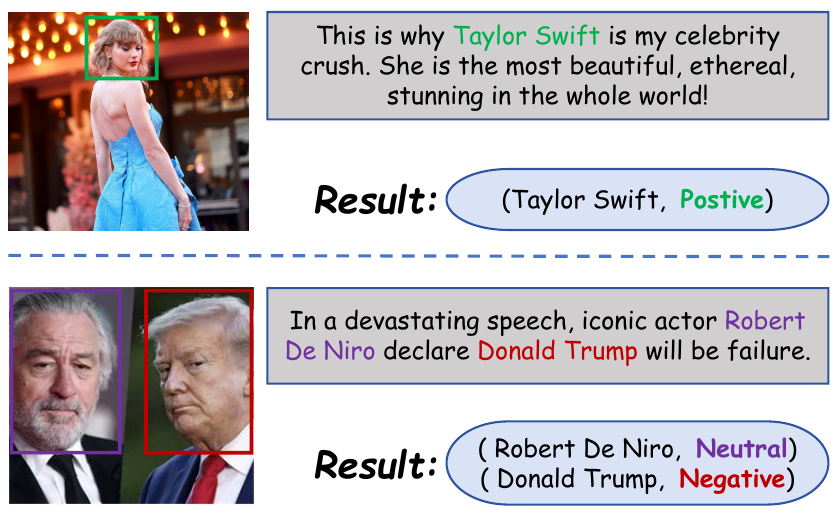}
        \caption{Example of MABSA. MABSA task utilizes both images and text to obtain aspects and their corresponding sentiment orientations.
        The color green represents positive, red indicates negative, and purple is neutral.}
        \label{fig:intro}
    \end{figure}

To assess the emotional orientation in human expression, traditional sentiment analysis methods need to adapt to this multimodal scenario, which is referred to as Multimodal Aspect-Based Sentiment Analysis (MABSA). It plays an important role in several applications, such as healthcare~\cite{health} and human-computer interaction~\cite{humanInteraction}. Specifically, MABSA aims to both extract aspect terms from text~\cite{wang2021} and assign sentiment label to each of these aspect terms~\cite{Ho2022, Zhang_Wang_Zhang_2021}. For example, as depicted in Fig. \ref{fig:intro}, MABSA aims to extract ``Taylor Alison Swift" and the corresponding sentiment is ``positive" from multimodal information including text and image.
    
With the significant improvement in the performance of large language model (LLM), LLM is attracting increasing attention. Despite the good performance of LLM in general tasks, such as image captioning~\cite{TCSVT_IC} and visual question-answering (VQA)~\cite{blip2, TCSVT_VQA}, it still performs limited in some complex scenarios~\cite{SASurvey}, leaving more uncertainties. Recently, researchers have started to pay attention to the adaptability of LLM in complex tasks, such as multimodal entity linking~\cite{GEMEL}, multimodal contextual object detection~\cite{zang2023contextual}, and machine translation with cultural awareness~\cite{yao2023empowering}.
However, regarding MABSA, the adaptability of LLM remains uncertain, and there is no established benchmark for comparison. Therefore, we seek to explore the suitability of LLM for MABSA, rethinking their role and necessity in this context, and introduce a basic benchmark for LLM-based methods in the MABSA task.
\textit{To the best of our knowledge, we are the first to explore the use of LLM for MABSA.} 
% In selecting the foundational large language models, we select Llama2 and ChatGPT for exploration. However, the poor experimental results indicate that LLMs do not perform well in the complex MABSA task. 

For our exploration, we design the LLM For Sentiment Analysis (LLM4SA) framework to evaluate the suitability of large language models (LLMs) for the Multimodal Aspect-Based Sentiment Analysis (MABSA) task. The framework leverages multimodal examples for in-context learning, where text and visual features are jointly processed to extract aspect terms and their corresponding sentiments. Specifically, we select well-established LLMs, including Llama2~\cite{llama2}, ChatGPT~\cite{GPT4}, and LLaVA~\cite{llava}, as the primary models for evaluation. For image feature extraction, we adopt the Llava pipeline, using a pre-trained vision transformer (ViT) to encode visual embeddings, which are then aligned with textual features through a projector for integration into the LLM.

In this framework, Llama2~\cite{llama2}, Llava~\cite{llava} and ChatGPT~\cite{GPT4} process structured inputs comprising visual tokens and text. These inputs include aspect terms and their sentiment labels (positive, negative, or neutral), with in-context learning (ICL) examples provided from the training set to guide the model’s predictions. The number of ICL examples ($\lambda$) is set to 10 for Llama2 and 5 for ChatGPT, considering their respective capabilities and constraints. We evaluate performance directly on test datasets and benchmark the results against traditional supervised learning methods.

Our results show that while LLMs demonstrate potential for multimodal understanding, they face significant challenges in addressing the fine-grained and complex requirements of MABSA. Additionally, their computational cost is significantly higher compared to SLM-based methods, which limits their practicality in real-world applications. These findings highlight the current limitations of LLMs and emphasize the need for further optimization to improve their adaptability to intricate multimodal sentiment analysis tasks.

\section{Related Work}
With the gradual evolution of large-scale models, an increasing number of universal large models have been developed, which are versatile for various downstream tasks such as ChatGPT\cite{GPT4}, Llama~\cite{llama}, and others~\cite{Vicuna}. These large models, by adjusting prompts, can handle tasks like text classification~\cite{textclassification} and machine translation~\cite{mt}.
However, the modern landscape of social media has introduced more intricate scenarios, necessitating the integration of multimodal processing for diverse tasks. This leads to the development of more multimodal large models like Flamingo~\cite{Flamingo} and etc., designed for tasks such as Visual Question Answering (VQA)~\cite{vqa} and Image Captioning~\cite{IC}.

Although multimodal large models are capable of addressing the complexity of multimodal scenarios~\cite{zhao2023mcl, zhao2021enhancing}, they still exhibit limitations when confronted with other intricate tasks. Research~\cite{SASurvey} has indicated that Large Language Model (LLM) perform inadequately in sentiment analysis task, falling short of traditional pretrained models~\cite{Roberta, DeBERTa, BART}. Additionally, researchers have noted the adaptation of LLM to various complex tasks.
For instance, \citet{LLMForTrans} have acknowledged that recent in-context learning utilizes lightweight prompts~\cite{lester2021prompt} to guide LLM for machine translation. However, it remains unclear whether this approach effectively injects cultural awareness into machine translation. To address this, a new prompt strategy and the process of constructing a culturally relevant parallel corpus have been proposed.
In the context of multimodal entity linking task~\cite{adjali2020multimodal, gan2021multimodal}, LLM has not received adequate attention and related solutions. To address this issue, \citet{GEMEL} propose a generative multimodal entity linking framework based on LLM, which directly generates target entity names.

Increasingly, researchers are not only focused on the versatility of large models themselves but also on the specificity of tasks in complex scenarios. Consequently, they are proposing relevant solutions to meet the demands of these diverse tasks.

\begin{figure*}[h]
    \centering
    \includegraphics[width=0.7\textwidth]{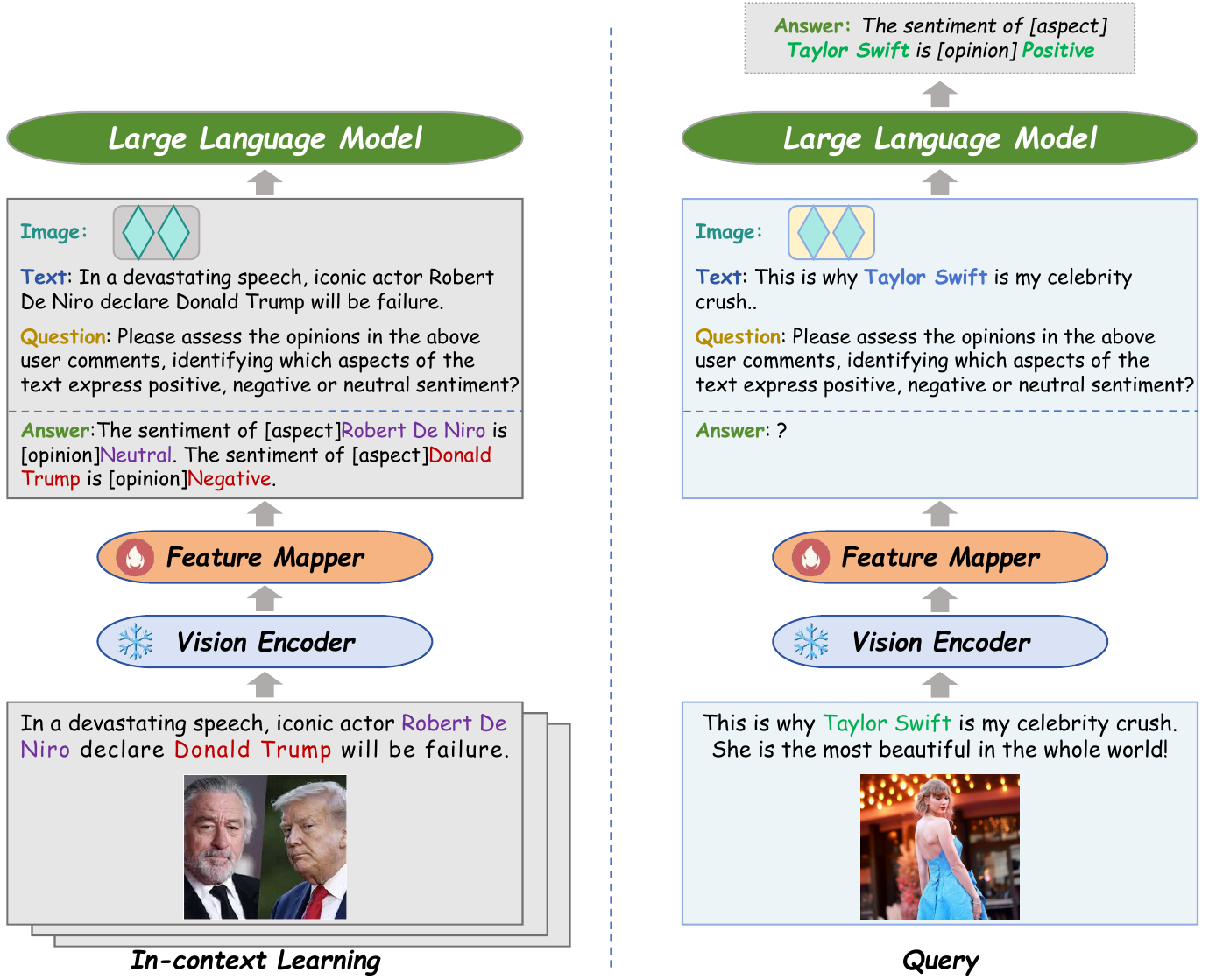}
    \caption{The framework of our LLM for multimodal aspect-based sentiment analysis (LLM4SA) method. The left side illustrates the in-context learning process, while the right side shows the user's input question. The image tensor is a vector obtained by encoding an image using ViT.}
    \label{fig:LLMFramework}
\end{figure*}

\section{Exploration of LLM for MABSA}
\label{Exploration}  
Large Language Models (LLMs) \cite{SASurvey} have drawn substantial attention within the AI community due to their remarkable ability to comprehend, reason with, and generate human language. Researchers leverage the formidable capabilities of LLM as the core components to tackle a variety of multimodal tasks. Thus, for comparison and giving insight into the application of LLM on complicated downstream tasks such as MABSA, we introduce our LLM For Sentiment Analysis (LLM4SA) framework. 

As shown in Fig. \ref{fig:LLMFramework}, our LLM4SA takes the multimodal examples for in-context learning and generates the aspect and sentiment. We employ Llama2-7b \cite{llama} as our default LLM.
In addition, we employ LLaVA~\cite{llava} and GPT3.5-turbo \cite{GPT4} for MABSA and evaluate their performance. The performance comparison of the LLM-based method and traditional method including textual and multimodal methods is shown in Table \ref{tab:LLMResult}.
For Llama and ChatGPT, performance assessments are conducted directly on the test dataset. Notably, In-context learning (ICL) examples are randomly selected from the training dataset.

\subsection{Visual Feature}
    To achieve feature alignment, we initially extract image features from the pre-trained vision encoder, which are then mapped into the text embedding space using a projector. These mapped features are subsequently input into the LLM as a visual prefix.
    As for image features, we adopt the image processing approach from Llava \cite{llava}. To obtain visual features from an input image $I$ corresponding to the text $t$, we utilize a pretrained visual model, ViT, which generates visual embeddings $Z_v \in R^{d_v}$, where $d_v$ represents the hidden state dimension of the vision encoder. The weights of the vision encoder remain frozen.
    
    To enable cross-modal alignment and fusion, we employ a projector $W^l$ to transform visual features into a soft prompt, effectively creating a visual prefix for the LLM input. In more detail, we use a linear layer to project image features, resulting in visual tokens $H_v$. Consequently, these visual tokens $H_v$ have the same dimensionality as the word embedding space in the language model. The calculation of visual tokens $H_v$ is as follows:
    \begin{align}
        Z_v &= ViT(I)\\
        H_v &= W^l Z_v
    \end{align}

    \input{table/llm}

\subsection{Large Language Models}
    \subsubsection{Llama2 and LLaVA for MABSA}
    We structure the visual tokens $H_v$ and text $t$ in the following format for input into the LLM. \textit{aspect} is a substring within the text $t$ and \textit{sentiment} is chosen from \textit{\{positive, negative, neutral\}}. 
    For example, in the text ``This is why Taylor Alison Swift is my celebrity crush. She is the most beautiful, ethereal, stunning in the whole world!" with the corresponding image, the \textit{aspect} is \textit{Taylor Alison Swift} and \textit{sentiment} is \textit{positive}.
    \textit{ICLs} represent $\lambda$ randomly selected instances from the training dataset for in-context learning, separated by the delimiter `[SEP]'. We set $\lambda$ to be 10. In ICLs, the content of the `[Answer]' field has already been completed based on annotations from the training dataset. In contrast, in the \textit{Query}, the `[Answer]' field is missing and needs to be generated by LLM. Finally, the evaluation of model performance is based on the content of the generated `[Answer]' field of Query.

    \subsubsection{ChatGPT for MABSA}
    We select GPT-3.5-turbo for testing. Due to the limitation of GPT, the number of In-context learning examples $\lambda$ is set to be 5. We utilize the completion capabilities of GPT's chat model to accomplish the task. The input text is structured as follows:
    ``Here is examples for MABSA: \textit{ICLs}. Based on the above examples, complete the following completion tasks: \textit{Query} "

% \subsection{Limited Speed}
%     We analyze the inference speed of various methods and present their speed in Table \ref{tab:LLMResult}. 
%     Our observation indicates that LLMs operate considerably slower than SLMs due to their increased parameters, extended input contexts, and additional response decay (if external APIs are utilized).

\section{Experiment}

\input{table/dataset}

\subsection{Datasets and Evaluation Metrics}
    To evaluate the performance of the dual-encoder transformer with cross-modal alignment, two MABSA benchmark datasets are used, mainly consisting of reviews on Twitter, including text and image. These datasets are Twitter-2015 and Twitter-2017, originally provided by \citet{TwitterPropose} for multimodal named entity recognition and annotated with the sentiment polarity for each aspect by \citet{TwitterLabel}. 
    
    An aspect is regarded as correctly predicted only if the aspect term and polarity respectively match the ground-truth aspect term and corresponding polarity. Table \ref{tab:dataset} summarizes the statistical characteristics of these two datasets. Precision, recall, and micro F1-score are used as evaluation metrics for MABSA. 
    Specifically, 
    precision refers to the proportion of true positive predictions out of all positive predictions made by the model. In other words, precision measures how many of the model's positive predictions are accurate.
    Recall is the proportion of true positive predictions out of all actual positive instances. Recall represents the ability to correctly identify positive instances, showing how many positives it managed to detect.
    Micro F1 score is the weighted harmonic mean of precision and recall. This score provides an overall assessment of the model's performance. The F1 score considers a balance between precision and recall, offering a single measure that combines both metrics and is robust for imbalanced class data.
    
\subsection{Implement Details}  
    In the selection of large-scale models, we chose Llama2 and LLaVA with 7B parameters and GPT3.5-turbo as the test models.
    For all experiments, the weights of DeBERTa and ViT are respectively initialized from pretrained DeBERTa-base and Vit-base-patch16-224-in21k. The hidden size $d$ is 768, the number of heads in cross-modal self-attention is 8, patch size $P$ is 14, MLP size is 3,072 and the number of attention heads is 12.
    Besides, AdamW optimizer \cite{AdamW} with a base learning rate of $2e^{-5}$ and warmup decay of 0.1 is used to update all trainable parameters. The maximum length and batch size are respectively set to 60 and 8. For training epochs, we leveraged an early stopping strategy with a patience of 3 to avoid overfitting.
    The Pytorch version used in all experiments in this article is 1.10. All experiments are conducted on RTX3090.

\subsection{Baselines}

To conduct a thorough comparison, we select multiple representative advanced methods based on SLM and LLM.  

\begin{itemize}
    \item RoBERTa \cite{Roberta} is an enhanced text encoding model built upon BERT by removing the next sentence prediction objective, training on longer sequences, and dynamically altering the masking pattern applied to the training data .
    \item DTCA \cite{DTCA} introduces a dual-encoder transformer, and improves cross-modal alignment by introducing auxiliary tasks and an unsupervised approach to enhance text and image representation. We reproduce the experiment and report the reproduction results.
    \item AoM~\cite{AoM} is an Aspect-oriented Method (AoM) addressing noise in MABSA. AoM introduces an aspect-aware attention module to detect aspect-relevant information from text-image pairs, integrating sentiment embedding and graph convolutional networks for precise sentiment aggregation.
    \item DQPSA~\cite{DQPSA} enhances multimodal aspect-based sentiment analysis by introducing a Prompt as Dual Query (PDQ) module for improved visual-language alignment and an Energy-based Pairwise Expert (EPE) module for span prediction, achieving notable performance improvements.
\end{itemize}

\begin{itemize}
    \item Llama2~\cite{llama2} is a large language model developed by Meta, featuring an optimized transformer architecture and advanced fine-tuning techniques. It is widely recognized for its robust performance across various natural language processing tasks, making it a reliable baseline for evaluating multi-modal sentiment analysis.
    \item Llava~\cite{llava} is a vision-language model designed for multi-modal tasks by integrating textual and visual modalities. Its dual-stream architecture and feature alignment strategies make it a competitive baseline for sentiment analysis tasks involving text and images.
    \item ChatGPT~\cite{GPT4} is a conversational AI model developed by OpenAI, designed for complex multi-turn dialogue and natural language understanding. With its broad training corpus and refined reasoning capabilities, it serves as an effective baseline for text-based sentiment analysis.
\end{itemize}

\input{table/LLMResult}
\subsection{Result and Discussion}
    % \begin{figure}[h]
    %     \centering
    %     \includegraphics[width=0.7\linewidth]{utils/LLMResult.png}
    %     \caption{Comparison with LLM Result on Twitter15}
    %     \label{fig:LLMResult}
    % \end{figure}
    % As depicted in the figure \ref{fig:LLMResult}, it is evident that our approach (TAM) exhibits a significant advantage in performance compared to methods based on large language model (LLM). It is worth emphasizing that even though approaches based on LLM is few-shot, they still demonstrate comparative ability.

    We wonder whether LLMs can outperform supervised SLMs in MABSA task purely through in-context learning.
    To this end, we select two methods based on pre-trained model for comparison on two datsets~\cite{TwitterDataset}, including the text-based RoBERTa method~\cite{Roberta} and the multimodal DTCA method~\cite{DTCA}. The performance of the DTCA is a result of our replication. As illustrated in Table \ref{tab:LLMResult}, from the experimental results, it is evident that LLM-based methods still exhibit a performance gap compared to traditional methods.
    We dive deep into the above results and analyze why LLMs fail to achieve satisfactory performance: 
    
    \textbf{Insufficient familiarity with downstream task specifics.}
    % MABSA是更细粒度的情感判断任务。对LLM来说，完成MABSA，需要先判断一句中存在多少个aspect-sentiment pairs，在判断句子中aspect和相应的sentiment。一个正确的prediction需要正确完成三个子任务。这就为LLM带来了很大的压力。
    As stated in \citet{ma2023large}, MABSA task is scarce in the widely-used instruction tuning datasets like \citet{wei2021finetuned} and \citet{wang2022super}.  Therefore it is likely that instruction-tuned LLMs are not well-acquainted with such MABSA task formats.
    However, MABSA is an intricate task in sentiment judgment. Without sufficient understanding of MABSA, LLM would find it challenging to accomplish this. Accomplishing the MABSA task necessitates initially determining the number of aspect-sentiment pairs within a sentence, followed by identifying the aspects within the sentence and subsequently recognizing the corresponding sentiment. Achieving a correct prediction requires successful completion of these three subtasks, which bring a significant burden on the LLM.

    \textbf{Limited number and effectiveness of samples.}
    % 受限于LLM推理速度和模型大小，ICL的数量不会设置过高。在有限的ICL数量中，LLM的学习就依赖于ICL中sample的有效性了。当sample不够具有代表性时，可能LLM并不能学习到对MABSA任务有用的信息。
    Due to limitations in LLM's reasoning speed and model size, the quantity of ICL (In-context Learning) will not be set excessively high. Within this limited number of ICL samples, the learning of LLM relies on the effectiveness of the sampled content.  In cases where the samples are not sufficiently representative, the LLM may fail to acquire valuable information relevant to the MABSA task.
    Meanwhile, SLMs can continually learn from more samples through supervised learning, widening the performance gap as annotated samples increase.

    \textbf{High time cost}
    As shown in Table \ref{tab:LLMResult}, LLMs operate considerably slower than SLMs due to their increased parameters, extended input contexts, and additional response decay (if external APIs are utilized)

\section{Conclusion}
This paper investigates the suitability of large language models (LLMs) for the Multimodal Aspect-Based Sentiment Analysis task, comparing them with supervised learning methods (SLMs) on two public datasets. Our findings reveal that LLMs, despite their strong capabilities in general multimodal tasks, face significant challenges in addressing the complex and fine-grained requirements of MABSA. Specifically, LLMs exhibit limitations in three key areas: insufficient familiarity with downstream task specifics, restricted learning from in-context examples due to sample representativeness and quantity constraints, and high computational time costs compared to SLMs.
These results highlight the current performance gap between LLM-based methods and traditional supervised methods in MABSA. While LLMs show potential for multimodal understanding, their effectiveness in tasks requiring intricate reasoning, such as MABSA, remains limited. Future research should focus on improving task-specific instruction tuning, enhancing sample effectiveness for in-context learning, and optimizing computational efficiency to better adapt LLMs for fine-grained multimodal sentiment analysis.

\bibliographystyle{ACM-Reference-Format}
\bibliography{sample-base}

\end{document}

%% file: table/llm.tex
\begin{table}[htbp]
    \centering
    \caption{The example of In-context learning (ICL) and query for LLM.}
    \begin{tabular}{cp{6cm}}
    \toprule
      & \multicolumn{1}{c}{Text} \\
    \midrule
    ICL  & [Image]\textit{\textcolor{blue}{$H_v$}}, [Text]\textit{\textcolor{blue}{$t$}}, [Question]Assess the opinions in the above user comments, identifying which aspects of the text express positive, negative or neutral sentiment.  [Answer]the opinion of [aspect]\textit{\textcolor{blue}{aspect}} is [sentiment]\textit{\textcolor{blue}{sentiment}}.  \\
    Query  & [Image]\textit{\textcolor{blue}{$H_v$}}, [Text]\textit{\textcolor{blue}{$t$}}, [Question]Assess the opinions in the above user comments, identifying which aspects of the text express positive, negative or neutral sentiment.  [Answer] \textbf{\textcolor{blue}{?}}.  \\
    \bottomrule
    \end{tabular}
\label{llm_input}
\end{table}

%% file: table/dataset.tex
\begin{table*}[htbp]
\centering
\caption{Statistics of datasets \cite{TwitterDataset} (\#S, \#A, \#Pos, \#Neu, \#Neg, MA, MS, Mean and Max denote numbers of sentences, aspects, positive aspects, neural aspects, positive aspects, multi aspects in each sentence, multi sentiments in each sentence, mean length and max length).}
\begin{tabular}{ccccccccccc}
\toprule
\multicolumn{2}{c}{\textbf{Datasets}}                       & \textbf{\#S}         & \textbf{\#A}         & \textbf{\#Pos}       & \textbf{\#Neu}       & \textbf{\$Neg}       & \textbf{MA}          & \textbf{MS}          & \textbf{Mean}        & \textbf{Max}         \\
\midrule
\multirow{3}{*}{\textbf{Twitter15}} & Train                & 2100                 & 3179                 & 928                  & 1883                 & 368                  & 800                  & 278                  & 15                   & 35                   \\
                                     & Dev                  & 737                  & 1122                 & 303                  & 670                  & 149                  & 286                  & 119                  & 16                   & 40                   \\
                                     & Test                 & 674                  & 1037                 & 317                  & 607                  & 113                  & 258                  & 104                  & 16                   & 37                   \\
\midrule
\multirow{3}{*}{\textbf{Twitter17}} & Train                & 1745                 & 3562                 & 1508                 & 1638                 & 416                  & 1159                 & 733                  & 15                   & 39                   \\
                                     & Dev                  & 577                  & 1176                 & 515                  & 517                  & 144                  & 375                  & 242                  & 16                   & 31                   \\
                                     & Test                 & 587                  & 1234                 & 493                  & 573                  & 168                  & 399                  & 263                  & 15                   & 38                   \\
\bottomrule
\end{tabular}
\label{tab:dataset}
\end{table*}

%% file: table/LLMResult.tex
% Table generated by Excel2LaTeX from sheet 'Sheet1'
% \begin{table}[htbp]
% % \renewcommand\arraystretch{1.3}
%   \centering
%   \caption{The result comparison between LLM-based method and traditional method. Mutli represents the multimodal method. and * represents the reproduction result.}
%     \begin{tabular}{c|l|ccc|ccc}
%     \toprule
%     \multirow{2}[2]{*}{\textbf{Type}} & \multicolumn{1}{c|}{\multirow{2}[2]{*}{\textbf{Approaches}}} & \multicolumn{3}{c|}{\textbf{Twitter15}} & \multicolumn{3}{c}{\textbf{Twitter17}} \\
%           &       & \textbf{F} & \textbf{P} & \textbf{R} & \textbf{F} & \textbf{P} & \textbf{R} \\
%     \midrule
%     \textbf{Text} & RoBERTa & 63.3  & 62.9  & 63.7  & 65.6  & 65.1  & 66.2  \\
%     \textbf{Multi} & DTCA* & 67.5  & 66.0  & 68.9  & 69.0  & 68.4  & 69.5  \\
%     \midrule
%     \multirow{2}[1]{*}{\textbf{LLM}} & Llama2 & 54.3  & 53.6  & 55.0  & 58.2  & 57.6  & 58.8  \\
%           & ChatGPT & 51.4  & 50.9  & 51.9  & 55.8  & 55.6  & 56.1  \\
%     \bottomrule
%     \end{tabular}%
%   \label{tab:LLMResult}%
% \end{table}%

% Table generated by Excel2LaTeX from sheet 'Sheet1'
\begin{table*}[htbp]
  \centering
  \caption{The result comparison between LLM-based method and traditional method. * represents the reproduction result. Time is inference seconds over 500 samples (run on single A100 GPU). The inference time for ChatGPT includes API calls and waiting time for inference.}
    \begin{tabular}{c|l|ccc|ccc|c}
    \toprule
    \multirow{2}[2]{*}{} & \multicolumn{1}{c|}{\multirow{2}[2]{*}{\textbf{Methods}}} & \multicolumn{3}{c|}{\textbf{Twitter15}} & \multicolumn{3}{c|}{\textbf{Twitter17}} & \multirow{2}[2]{*}{\textbf{Time(/s)}} \\
          &       & \textbf{F1} & \textbf{Precision} & \textbf{Recall} & \textbf{F1} & \textbf{Precision} & \textbf{Recall} &  \\
    \midrule
    \textbf{Text} & RoBERTa~\cite{Roberta} & 63.30  & 62.90  & 63.70  & 65.60  & 65.10  & 66.20  & 3.18  \\
    \midrule
    \multirow{3}[2]{*}{\textbf{Multimodal}} & DTCA*~\cite{DTCA} & 67.50  & 66.00  & 68.90  & 68.97  & 68.42  & 69.53  & 9.21  \\
          & AoM~\cite{AoM}   & 68.60  & 67.90  & 69.30  & 69.70  & 68.40  & 71.00  & 16.47  \\
          & DQPSA~\cite{DQPSA} & \textbf{71.90 } & \textbf{71.70 } & \textbf{72.00 } & \textbf{70.60 } & \textbf{71.10 } & \textbf{70.20 } & 10.24  \\
    \midrule
    \multirow{3}[2]{*}{\textbf{LLM}} & Llama2~\cite{llama2} & 54.29  & 53.60  & 55.00  & 58.19  & 57.60  & 58.80  & 1214.11  \\
          & LLaVA~\cite{llava} & 55.62  & 56.70  & 54.57  & 61.74  & 62.40  & 61.08  & 888.57  \\
          & ChatGPT\cite{GPT4} & 51.40  & 50.90  & 51.90  & 55.80  & 55.60  & 56.10  & 5643.74  \\
    \bottomrule
    \end{tabular}%
  \label{tab:LLMResult}%
\end{table*}%